\documentclass[a4paper,conference]{IEEEtran}
\usepackage{ifpdf}
\usepackage[normalem]{ulem}
\usepackage{multirow}
\useunder{\uline}{\ul}{}

\usepackage{cite}

\ifCLASSINFOpdf
  \usepackage[pdftex]{graphicx}
\else
\fi

\usepackage{amsmath}

\usepackage{algorithmic}

\usepackage{array}

\ifCLASSOPTIONcompsoc
 \usepackage[caption=false,font=normalsize,labelfont=sf,textfont=sf]{subfig}
\else
 \usepackage[caption=false,font=footnotesize]{subfig}
\fi

\usepackage{dblfloatfix}

\usepackage{url}

\begin{document}
%
\title{Class2Str: End to End Latent Hierarchy Learning}

\author{\IEEEauthorblockN{Soham Saha
\IEEEauthorrefmark{1}\IEEEauthorrefmark{2},
Girish Varma\IEEEauthorrefmark{1}\IEEEauthorrefmark{3} and C.V. Jawahar\IEEEauthorrefmark{4}}
\IEEEauthorblockA{\IEEEauthorrefmark{2}\IEEEauthorrefmark{3}\IEEEauthorrefmark{4}Centre for Visual Information Technology, KCIS,\\
International Institute of Information Technology, Hyderabad\\\IEEEauthorrefmark{2}Flipkart Internet Pvt. Ltd, Bengaluru, India\\
Email: \IEEEauthorrefmark{2}soham.saha@research.iiit.ac.in,
\IEEEauthorrefmark{3}girish.varma@iiit.ac.in,
\IEEEauthorrefmark{4}jawahar@iiit.ac.in}}

\vspace{-1em}


%


\maketitle

\begin{abstract}
Deep neural networks for image classification typically consists of a convolutional feature extractor followed by a fully connected \emph{classifier} network. The predicted and the ground truth labels are represented as one hot vectors. Such a representation assumes that all classes are equally dissimilar. However, classes have visual similarities and often form a hierarchy. Learning this latent hierarchy explicitly in the architecture could provide invaluable insights. We propose an alternate architecture to the classifier network called the Latent Hierarchy (LH) Classifier and an end to end learned Class2Str mapping which discovers a latent hierarchy of the classes. We show that for some of the best performing architectures on CIFAR and Imagenet datasets, the proposed replacement and training by LH classifier recovers the accuracy, with a fraction of the number of parameters in the classifier part. Compared to the previous work of HDCNN, which also learns a 2 level hierarchy, we are able to learn a hierarchy at an arbitrary number of levels as well as obtain an accuracy improvement on the Imagenet classification task over them. 
We also verify that many visually similar classes are grouped together, under the learnt hierarchy.
\end{abstract}


%
\IEEEpeerreviewmaketitle

\footnote{\IEEEauthorrefmark{1}Authors have equal contribution}
\vspace{-2em}
\section{Introduction}

\label{sec-intro}
Deep neural networks for image classification typically consist of a \emph{feature extractor} network with alternating convolutional and pooling layers \cite{AlexNet,Resnet,Inception}. The feature extractor is followed by a \emph{classifier network}, which maps the feature vectors to the class probabilities. This layer is typically a multi layered perceptron which is preferred since the predicted and the ground truth labels are represented as one hot vectors. The one hot representation assumes that the classes are independent of each other. However the classes typically are not completely independent. They have visual similarities and often form a hierarchy. For example the CIFAR 100 dataset has super classes, each of which consists of 20 subclasses. The Imagenet dataset  is organized according to the WordNet hierarchy \cite{ImageNet}. Even in datasets where a hierarchy is not mentioned explicitly, the classes might have some latent hierarchical structure. Learning this latent hierarchy explicitly, via a deep learning architecture could improve the performance of the model as it increases the ease of classification. Moreover, the fully connected classifier is known to have highly redundant parameters. Alternative well designed architectures for the classifier could reduce the parameter size too.

\begin{figure}[!tbh]
\centering
\includegraphics[width=7cm, height =3.5cm]{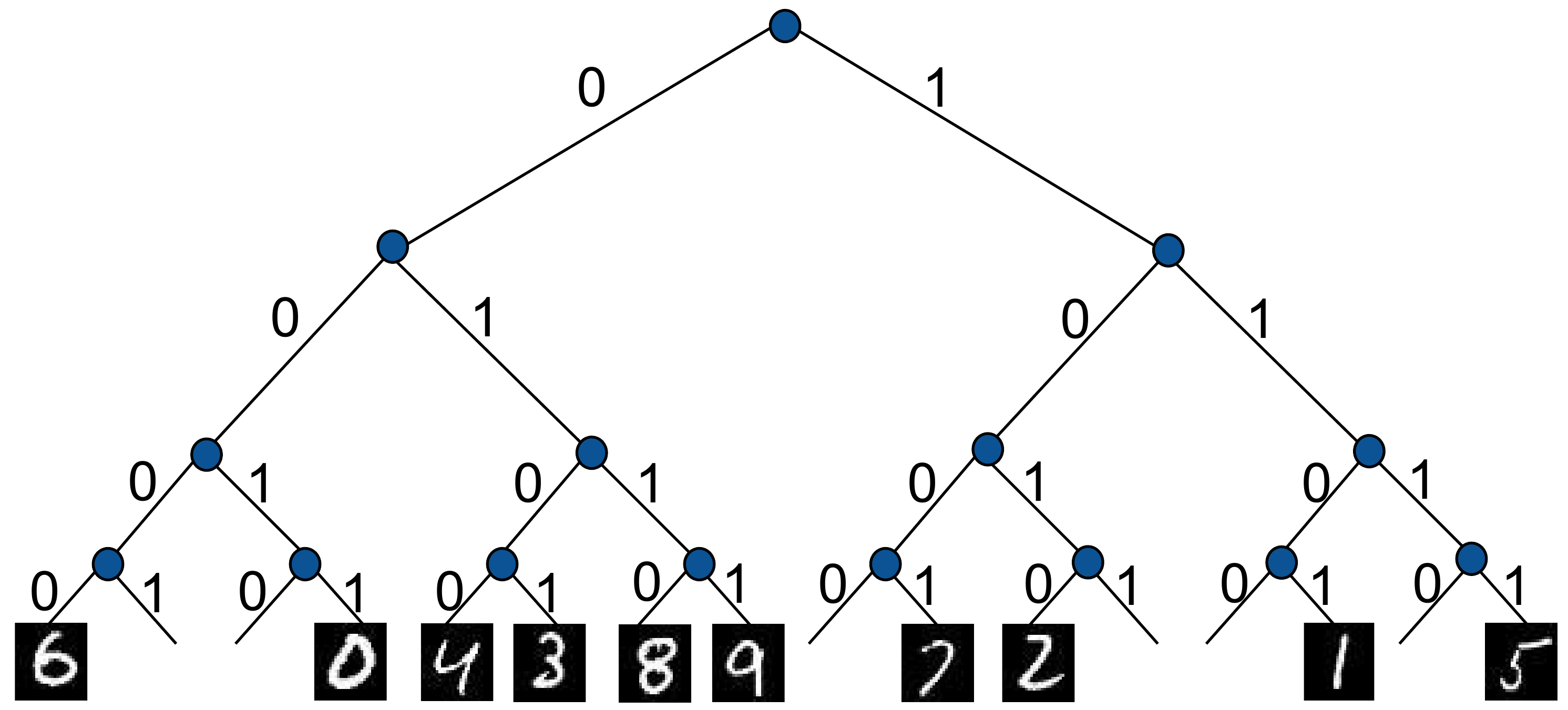}
\vspace{-0.8em}
\caption{ The hierarchy tree for MNIST learned by our proposed Class2Str network and Latent Hierarchy Classifier. Some of the visually similar digits such as 3,8 and 9 have a common prefix resulting in their close proximity in the tree. Also, 7 and 2 are in proximity on account of being visually similar.}
\label{fig:minst_tree}
\centering
\vspace{-1.5em}
\end{figure}

In this paper,  we propose to discover multi-level latent hierarchies in classes while maintaining image classification performance. We encode the latent hierarchy as a binary tree and map each leaf node to a binary string. We propose a deep learning architecture which consists of a Latent Hierarchy Classifier (replaces the Fully Connected classifier), Class2Str and Str2Class networks for discovering the hierarchy, represented as a string embedding (see Section \ref{sec-arch}). We identify the constraints that need to be satisfied by the model in order to achieve a high accuracy, and propose a structured loss function for learning the latent hierarchy (see Section \ref{sec-loss}).  We benchmark the performance of the proposed architecture and training methods, against some of the best performing architectures with classifier networks, in popular image classification datasets, and show comparable accuracies (see Section \ref{sec-results}). Compared to HD-CNN \cite{HDCNN} which also discovers a 2 level hierarchy, our method gives better accuracies using only a one third the number of parameter (See Table \ref{tab:4}) in the classifier network.  
We verify that the string embedding learnt corresponds to some latent hierarchy, by running the same experiments using a hard-coded random string embedding where we observe a decrease in accuracy (see Section \ref{sec-results}).

\section{Related Works}

\begin{figure*}
\centering
\includegraphics[width=15cm]{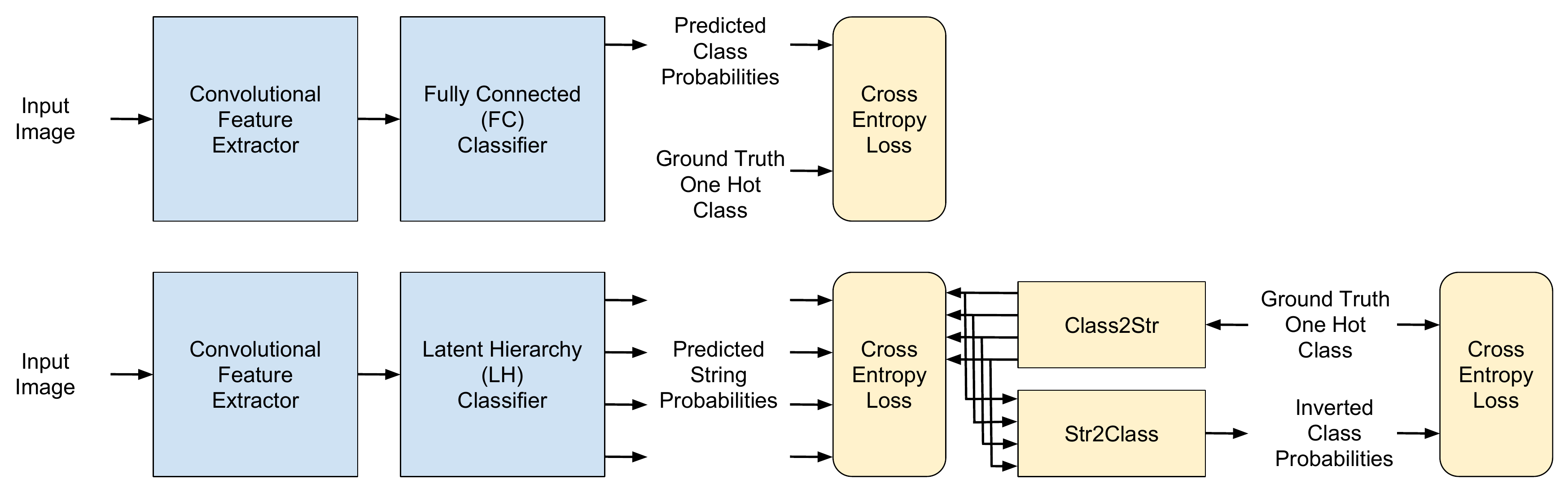}
\vspace{-1em}
\caption{The top half of the image shows a traditional classification architecture. The bottom half of the image shows the proposed architecture. The yellow blocks denote those parts which are used only during training phase.}
\label{fig:arc}
\vspace{-1.5em}
\end{figure*}

\textbf{Latent Hierarchy Learning.}
Though there is a vast literature in using class hierarchical structures in image classification (see \cite{HeirarchySurvey} for a survey), it is not very well studied using deep neural network based models. An early attempt at learning latent hierarchies was done by Srivastava et al. \cite{DiscTraLearnTree}. Their main goal is transferring knowledge between classes to improve the results with insufficient training examples. Deng et al. \cite{LabRelGraphs} uses prior knowledge about relations and hierarchies among classes to improve the classification accuracy. 

Yan et al. \cite{HDCNN} proposed a deep neural network model called HD-CNN to learn a latent hierarchy. They first train a fine grained classifier and then use the confusion matrix to identify some coarse grained categories. A separate fine grained classifier is trained for each of these coarse grained categories to improve the accuracy. Thus, HD-CNN discovers a 2 level hierarchy with a 2 stage training process, while we discover a multilevel hierarchy using a single stage end to end training. The comparison with HD-CNN is further enlisted in Table \ref{tab:4}. The Network of Experts \cite{NoE} model experiments with different strategies to group fine categories into coarse ones. However both these results have worked with a $2$ level hierarchy (the coarse and the fine categories). Our proposed method focuses on mutli-level hierarchies.

\textbf{Embeddings.}
Our methods resemble some of the embedding techniques that have been proposed previously. Target coding \cite{target_coding} suggests to replace the 1-hot representation of the labels by an error correcting code (Hadamard code). The mapping between the labels and the code words are fixed statically. However in our case, we are mapping labels to strings, which does not satisfy any non trivial minimum distance property. Moreover the mapping between labels to strings is learnt in an end to end fashion, unlike in their case. Our method is more inspired by Huffman’s coding which naturally gives a hierarchy rather than error correcting codes which has a minimum distance property. DeViSe \cite{devise}(and the follow up works related to semantic embedding and zero shot learning) suggests to replace the 1-hot vector representation of the labels by a word embedding which is learnt separately to model language (using text data). Their main goal is to use the word embedding to get semantic meaning of the labels. Hence the model is able to assign labels to images even if the dataset is small (zero-shot learning). Our goal is to learn a latent hierarchy among the classes that aids in classification. Latent hierarchy learning is ensured using a structured loss function, while they use similarity loss (for eg. cosine) between the label embeddings and image features. Our embedding to binary strings requires considerably less memory than their word embedding to real vectors (to be saved as a table). Also, they need to use a nearest neighbor search for finding the label, while we can do it by a simple binary tree traversal.

\section{Approach}
\label{sec-arch}
We propose an alternate architecture and training method for the classifier network, which can discover a latent hierarchy of the classes of arbitrary depth (see Figure \ref{fig:arc}). We take advantage of the fact that hierarchies can be represented by a binary tree with the leaf nodes being the classes (see Figure \ref{fig:minst_tree}). Any path in the tree from the root to a leaf can be represented as a unique binary string. Hence a binary tree defines a one to one mapping from classes to binary strings. Conversely, any one to one mapping from classes to binary strings could also be converted into a hierarchy, by considering the prefix tree of the strings (reminiscent of Huffman's codes from coding theory). Hence learning a hierarchy is equivalent to learning a one to one mapping from classes to strings (See Figure \ref{fig:minst_tree}).  The proposed architecture consists of the following networks:

\begin{figure*}
\centering
\includegraphics[scale=0.45]{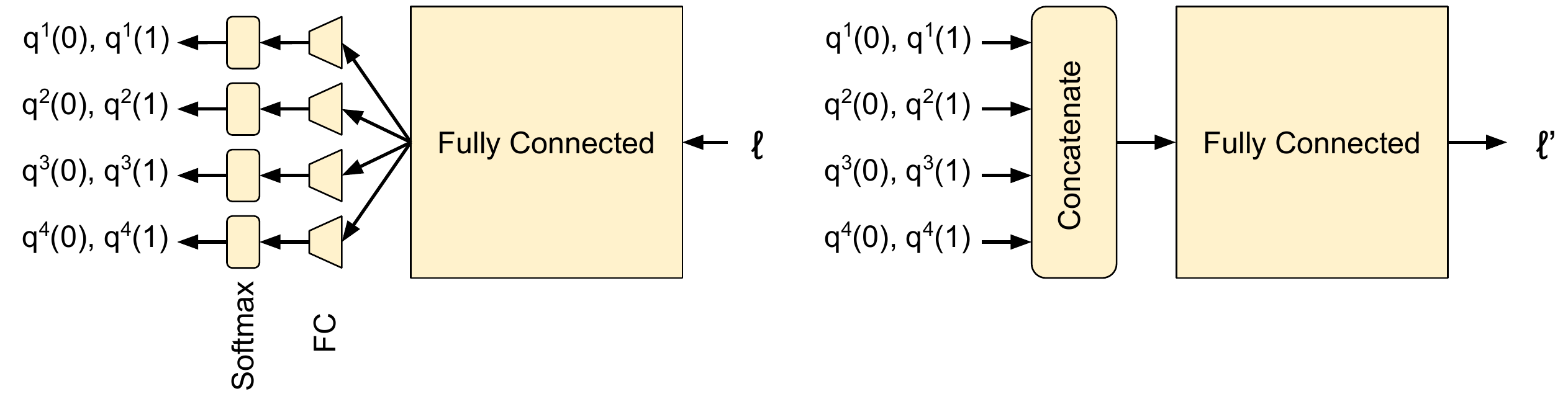}
\includegraphics[scale=0.45]{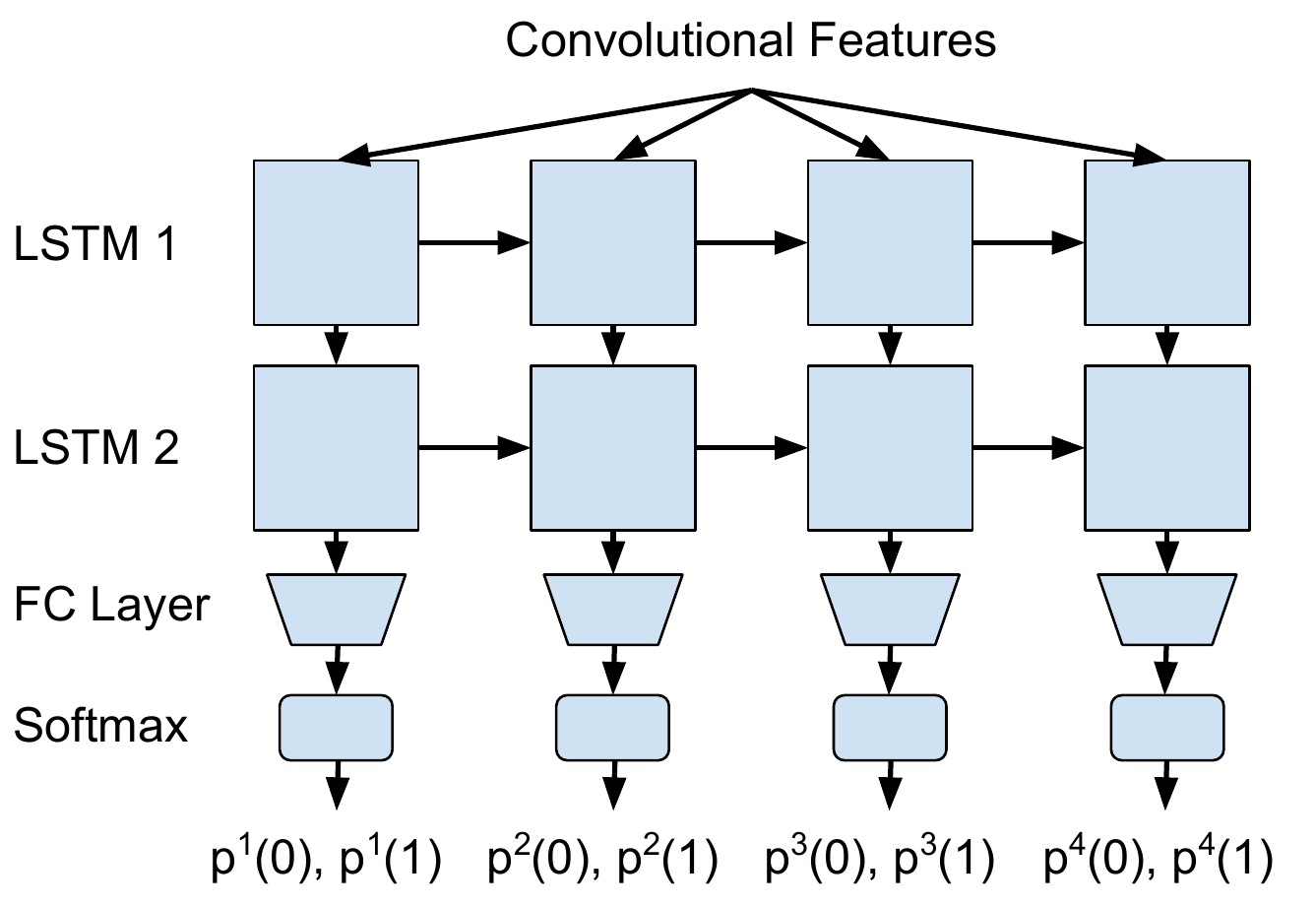}
\vspace{-1em}
\caption{[Yellow] The architectures of the proposed Class2Str (left) and Str2Class (middle) networks. Class2Str converts the label space representation into a latent binary string representation while Str2Class does vice-versa. [Blue] The architecture of the proposed LH Classifier which takes as input the convolutional features. This Classifier is plugged into the baseline network after removing the Fully Connected Layers. (right)}
\label{fig:c2s_rnn_classifier}
\vspace{-1.5em}
\end{figure*}

\textbf{Class2Str.}
The Class2Str network takes the one hot encoding of the ground truth class labels (denoted by $\ell$) and produces a probability distribution of strings $\in \{0,1\}^L$. The probability distribution is given by $2L$ variables, $\left(q^1(0), q^1(1)\right),\cdots , \left(q^L(0), q^L(1)\right)$ where $q^i(b)$ denotes the probability of the $i$th bit to be $b$. We obtain the string encoding by taking the maximum probability value in every bit. This is defined by the function:
\begin{equation}\label{eqn:s}
s(q) = a_1, a_2, \cdots a_L \text{ where } a_i = \text{argmax}_{b \in \{0,1\}}~ q^i(b).
\end{equation}
We experimented with a fully connected layer for this network. The probabilities for each bit in the string ( $q^i(0),q^i(1)$) by using a sequence of fully connected layers which maps the latent space to 2 dimensions and applying a softmax (see Figure \ref{fig:c2s_rnn_classifier}). During testing phase, we can replace this network by a look up table with entries corresponding to the string encoding for each class (see Section \ref{sec-exp}, for the method of operation during testing phase).


\textbf{Latent Hierarchical (LH) Classifier.} 
The fully connected classifier in the traditional architectures, is designed to predict class probability scores. The Class2Str network replaces the classes with strings. Hence we need the classifier network to predict a probability distribution over strings given by $2L$ variables. We will denote the probability distribution which is the output of the Class2Str network by $\left(p^1(0), p^1(1)\right),\cdots , \left(p^L(0), p^L(1)\right)$, where $p^i(b)$ denotes the probability that the $i$th bit is $b \in \{0,1\}$. Each bit of the predicted string, is supposed to be a binary classifier which given the previous bits, tries to best divide the data in the subclasses. With different prefixes, the classifier should be able classify according to a different criterion. This dependence on the previously predicted bits is essential to forming a hierarchy. Recurrent Neural Networks are best known to model these kind of dependencies. Hence we experiment with a multi layered unidirectional LSTM network \cite{Lstm}. The input is fed into a recurrent neural network repeated for $L$ time steps, where $L$ is the length of the string. We use single or double layered LSTM networks for the LH Classifier. The output of the LSTM network is passed through a fully connected layer which reduces the dimension to 2. Applying the softmax function, we obtain the probability scores for each bit of the string (see Figure \ref{fig:c2s_rnn_classifier}).


\textbf{Str2Class.}
Apart from these two, we have a third network called Str2Class, which is essential for ensuring that the string encoding learned by Class2Str is \emph{one to one}. The Str2Class network tries to decode the string produced by Class2Str to class probabilities (denoted by $\ell'$) which is close to $\ell$ (the one hot vectors of the ground truth labels). For Str2Class, we experiment with a fully connected network similar to the Class2Str network. We first concatenate the 2 dimensional probability vectors for each bit in to a $2L$ dimensional vector and pass it through a fully connected layer (see Figure \ref{fig:c2s_rnn_classifier}).
This network can also be removed during the testing phase (see Section \ref{sec-exp}, for the method of operation during testing phase).


\begin{table*}[!tbh]
\centering

\caption{The architectures of the networks used in our experiments. The sequence represents number of channels in each layer, the number of hidden units, the hidden state size of LSTM for the feature extractor, FC Classifier and LH Classifier respectively.}
\vspace{-0.6em}
\label{tab:data_arc}
\scalebox{1.05}{
\begin{tabular}{|c|c|c|c|c|c|}
\hline
\textbf{Dataset}                       & \textbf{Convolutional Feature Extractor}                                                                                                                  & \textbf{Fully Connected Classifier}            & \textbf{RNN Classifier} & \textbf{Class2Str} & \textbf{Str2Class} \\ \hline
MNIST                                  & 16 - pool - 32 - pool - 64                                                                                                                                & 3136-500-10                                    & 3136 - 10 - 10          & 10 - 500           & 500 - 10           \\ \hline
CIFAR 10                               & \begin{tabular}[c]{@{}c@{}}64 - 64 - pool - 128 - 128 - pool - \\ 256 - 256 - 256 - pool - 512 - 512 -\\ 512 - pool - 512 - 512 - 512 - pool\end{tabular} & 512 - 1024 - 1024 - 10                         & 512 - 20 - 20           & 10 - 500           & 500 - 10           \\ \hline
CIFAR 100                              & \begin{tabular}[c]{@{}c@{}}64 - 64 - pool - 128 - 128 - pool - \\ 256 - 256 - 256 - pool - 512 - 512 -\\ 512 - pool - 512 - 512 - 512 - pool\end{tabular} & 512 - 1024 - 1024 - 100                        & 512 - 40 - 40           & 100 - 1000         & 1000 - 100         \\ \hline
Imagenet 1K                             & \begin{tabular}[c]{@{}c@{}}64 - 64 - pool - 128 - 128 - pool - \\ 256 - 256 - 256 - pool - 512 - 512 -\\ 512 - pool - 512 - 512 - 512 - pool\end{tabular} & 25088 - 4096 - 4096 - 1000                     & 25088 - 2100 - 100      & 1000 - 2000        & 2000 - 1000        \\ \hline
\end{tabular}}
\vspace{-2em}
\end{table*}

\section{Training with Structured Loss}
\label{sec-loss}

Training the proposed architecture present some difficulties. One of the main problem is of ensuring that the \emph{Class2Str} encoding is one to one (ie. distinct classes are encoded as distinct strings). This is a discrete constraint and we need to design a continuous loss function that is minimized when this constraint is satisfied. Our solution consists of the \emph{Str2Class} network which inverts the Class2Str mapping. We design a continuous loss function which using the outputs of the Str2Class function, ensures that the Class2Str function is one to one on convergence by assigning appropriate weight to the specific component in our loss. Furthermore, we use a structured loss function (Equation \ref{eqn:struc-loss}), so that the string embedding gives rise to a latent hierarchy.

The loss function is defined by first identifying the constraints that need to be satisfied to get a high accuracy image classifier. Firstly, the Class2Str encoding must map distinct $\ell$ (classes)  to distinct $s(q)$'s. Without this, we cannot obtain a unique class label from the predicted string. This is satisfied if Str2Class is able to invert  the Class2Str mapping. That is $\ell$ must be equal to $\ell'$ (notation from the previous section). Secondly, the string encoding given by the Class2Str network much match the string predicted by the LH classifier. That is $s(p) = s(q)$ where $s$ is defined in Equation \eqref{eqn:s}. 

The first constraint can be modeled by a cross entropy term $H(\ell, \ell')$. For the second constraint, we first strengthen it, to say that the distribution $p, q$ are equal. Hence this can also be modeled by a cross entropy term $H(p,q) = \sum_{i=1}^L H(p^i,q^i)$. So a possible loss function is the following:
\begin{equation}\label{eqn:basic_loss}
\alpha H(\ell, \ell') + \beta \sum_{i=1}^L  H(p^i, q^i) 
\end{equation}
where $H$ denotes the cross entropy between distributions.

The above loss function is straightforward, but we encounter two problems when we train the network. On back-propagating the above loss function, $q^i(0), q^i(1)$ tends to converge towards the uniform distribution. When the $q^i$'s are close to unbiased, the string encoding function $s$ is not robust i.e a small change in the value of $q^i$ changes the string encoding. This results in low generalization accuracy.  Hence we add an additional constraint that $q^i$'s should be biased towards $0$ or $1$. We ensure this by using the following  regularizer which is maximized when these distributions are fully biased:
\begin{equation}\label{eqn:bias_reg}
\sum_{i=1}^L \left( q^i(0)^2 + q^i(1)^2 \right).
\end{equation}
 Equation \eqref{eqn:basic_loss} gives equal weight for each position of $p^i, q^i$. This implies that an error in the first as well as last position is equally penalized. However making an error at the top level of a hierarchy is in some sense greater than a error at the bottom level. An error at the first position, results in classification of the input to a class which is highly dissimilar to the ground truth. However strings having long common prefix, correspond to similar classes and the cost of error must be less. Hence we use the following loss term instead, where $\mu \in (0,1)$ is hyperparameter, to achieve better accuracies.
$\mu^i$ is evidently the most important hyperparameter which helps in learning of the correct hierarchy since this ensures that misclassification at the initial bits of string incur a larger penalty in the eventual loss. Hence this factor serves as a decay constant for the sequential penalty incurred in the subsequent bits of the predicted string. After careful tuning of this sensitive hyperparameter, we found that a decay factor of 0.8 for $\mu$ gives the best results.

\begin{equation}\label{eqn:struc-loss}
\sum_{i=1}^L \mu^i H(p^i, q^i)
\end{equation}

We also use a $L^2$ regularizer for all the weight in the network. Hence the final loss function that we minimize is the sum of the losses for each constraint, given bellow:
\begin{align}\label{eqn:loss}
\alpha H(\ell, \ell') + \beta \sum_{i=1}^L \mu^i H(p^i, q^i)  &- \gamma \sum_{i=1}^L \left( q^i(0)^2 + q^i(1)^2 \right) \nonumber \\ &+ \delta L^2(W) 
\end{align}

where $H$ denote the cross entropy function, $L^2(W)$ denotes the sum of squares of all the weights  and  $\alpha, \beta, \gamma,\delta$ are hyper parameters. Note that the term corresponding to Equation \eqref{eqn:bias_reg} is negative, since it needs to be maximized.

\begin{table}
\centering
\label{tab:4}
\caption{Comparison on parameter reduction and error rates with HD-CNN on the CIFAR-100 and Imagenet 1K dataset. Our method does better than HD-CNN in terms of number of parameters on CIFAR-100 and both in terms of error rate as well as number of parameters on Imagenet 1K.
}
\vspace{-0.5em}
\scalebox{0.85}{
\begin{tabular}{|c|c|c|c|}
\hline
{ \textbf{Network}} & { \textbf{Dataset}} & { \textbf{Error Rate}} & { \textbf{\# Parameters}} \\ \hline \hline
HD-CNN                 & CIFAR-100              & 32.62                     & $\sim$15M                    \\ \hline
\textbf{LHC (Ours)}    & CIFAR-100              & 35.33                     & \textbf{14.7M}               \\ \hline \hline
HD-CNN                 & Imagenet 1K            & 31.34                     & $\sim$220M                   \\ \hline
\textbf{LHC (Ours)}    & Imagenet 1K            & \textbf{29.89}            & \textbf{87.2M}               \\ \hline
\end{tabular}}
\vspace{-2em}
\end{table}

\begin{table*}
\centering
\caption{The accuracy and parameter reductions achieved by our proposed method on the respective datasets. Also, accuracies of some of the best performing models (Maxout \cite{maxout}, Network in Network \cite{NIN}, Deeply Supervised Networks \cite{deep_super_nets}) on the given datasets are shown. The third column shows the accuracy for an \textit{FC classifier with the same number of parameters as the LH Classifier} is used.}
\vspace{-0.5em}
\scalebox{1.2}{
\begin{tabular}{|c|c|c|c|c|c|c|c|}
\hline
Dataset                    & \begin{tabular}[c]{@{}l@{}}\% Acc of\\ FC\\ Classifier\end{tabular} & \begin{tabular}[c]{@{}l@{}}\% Acc of\\ LH\\ Classifier\end{tabular} & \begin{tabular}[c]{@{}l@{}}\% Acc of\\ reduced FC\\ Classifier\end{tabular} & \multicolumn{1}{c|}{\begin{tabular}[c]{@{}c@{}}\#parameters \\ in FC \\ Classifier\end{tabular}} & \multicolumn{1}{c|}{\begin{tabular}[c]{@{}c@{}}\#parameters\\  in \\ LH Classifier\end{tabular}} & \multicolumn{1}{c|}{\begin{tabular}[c]{@{}c@{}}Reduction\\ in \\ parameters\end{tabular}} & \multicolumn{1}{c|}{\begin{tabular}[c]{@{}c@{}}Reduction in\\ test time\\ per image\end{tabular}} \\ \hline
MNIST                      & 99.38                                                                     & 99.36                                                                    & 98.45                                                                            & 1.61 M                                                                                           & 31 K                                                                                             & 98\%                                                                                      & 13.9\%                                                                                            \\ \hline
\multirow{2}{*}{CIFAR 10}  & 90.43                                                                    & 90.51                                                                    & 88.40                                                                            & 1.58 M                                                                                           & 6 K                                                                                              & 99\%                                                                                      & 15.9\%                                                                                            \\ \cline{2-8} 
                           & \multicolumn{7}{l|}{\% Acc in Maxout: 90.65, Network in Network: 91.2, Deeply Supervised Networks : 91.78}                                                                                                                                                                                                                                                                                                                                                                                                                                                                                                                                                                      \\ \hline
\multirow{2}{*}{CIFAR 100} &  64.65                                                                   & 64.67                                                                    & 57.90                                                                            & 1.58 M                                                                                           & 30 K                                                                                             & 98\%                                                                                      & 14.8\%                                                                                            \\ \cline{2-8} 
                           & \multicolumn{7}{l|}{\% Acc in Maxout: 61.43, Network in Network: 64.32, Deeply Supervised Networks : 65.43}                                                                                                                                                                                                                                                                                                                                                                                                                                                                                                                                                                                                         \\ \hline
Imagenet 1K           &  70.51                                                                   & 70.11                                                                    &  67.88                                                                           & 123.63 M                                                                                         & 72.42 M                                                                                           & 41\%                                                                                      &  5.5\%                                                                                                \\ \hline
\end{tabular}}
\label{tab:2}

\end{table*}


\vspace{-0.3em}
\section{Dataset and Experiments}\label{sec-exp}
Our experiments were performed on MNIST, CIFAR10, CIFAR100 \cite{cifar} and Imagenet 1K \cite{Imagenet1k} datasets (ordered according to gradual increasing complexity). MNIST consists of single channel, ($28 \times 28$) images with $10$ classes. CIFAR10 images has $3$ channels with slightly bigger ($32 \times 32$) sizes, and $10$ classes. As a next step we moved to CIFAR100, which has $100$ classes but with the same image size. we followed this up by experimenting on  Imagenet 1K which consists of 1.28 million images with 1000 classes\cite{Imagenet10K}. For this dataset, We used color images of size $224\times 224$ randomly cropped from the rescaled images with the lower dimension being 256.

We build on the VGG16 architecture on Imagenet. For CIFAR (smaller image sizes), we use a varient of VGG16 and the LeNet architecture for MNIST. The exact details of the architectures for the experiments are provided in Table \ref{tab:data_arc}.

\textbf{Training Phase.}
We train a traditional network (for the image classification task), that consists of convolutional and fully connected layers (denoted as the Base Model) to convergence. Then, we replace the FC classifier with the proposed LH classifier, without altering the weights of the convolutional feature extractor. The LH classifier network is then trained along with the Class2Str and Str2Class networks by back-propagating the gradients calculated from the loss function (details mentioned in Section \ref{sec-loss}). The convolutional weights are not updated with the gradients during the backward pass. We use the Adam algorithm for the optimization. Hyperparameters $\alpha,\beta,\gamma,\delta$ are initially chosen so that each term in the loss function has the same order of magnitude. Furthermore $\gamma$ is reduced to as small a value as possible in order for the probabilities $q$ to remain highly biased.

\textbf{Testing Phase.}
During the testing phase the Class2Str and Str2Class networks are removed. Given an image, the LH classifier predicts a string $s(p)$. We maintain a look up table containing the one to one mappings of strings corresponding to each of the classes, learnt by the Class2Str network. The predicted class for the image is the the one corresponding to the string $s(p)$, in the look up table. The predicted strings for the test images are compared with the learnt strings for each class by the Class2Str network. If all bits match, we count it as a correct prediction.


\section{Results and Discussion}\label{sec-results}



We describe our results about the proposed model by first analyzing the latent hierarchy discovered by the model, then by comparing the accuracies achieved, and finally in terms of the parameters used in the classifier.

\begin{figure}[!tbh]
\vspace{-1em}
\centering
\includegraphics[height=6cm,width=11.5cm]{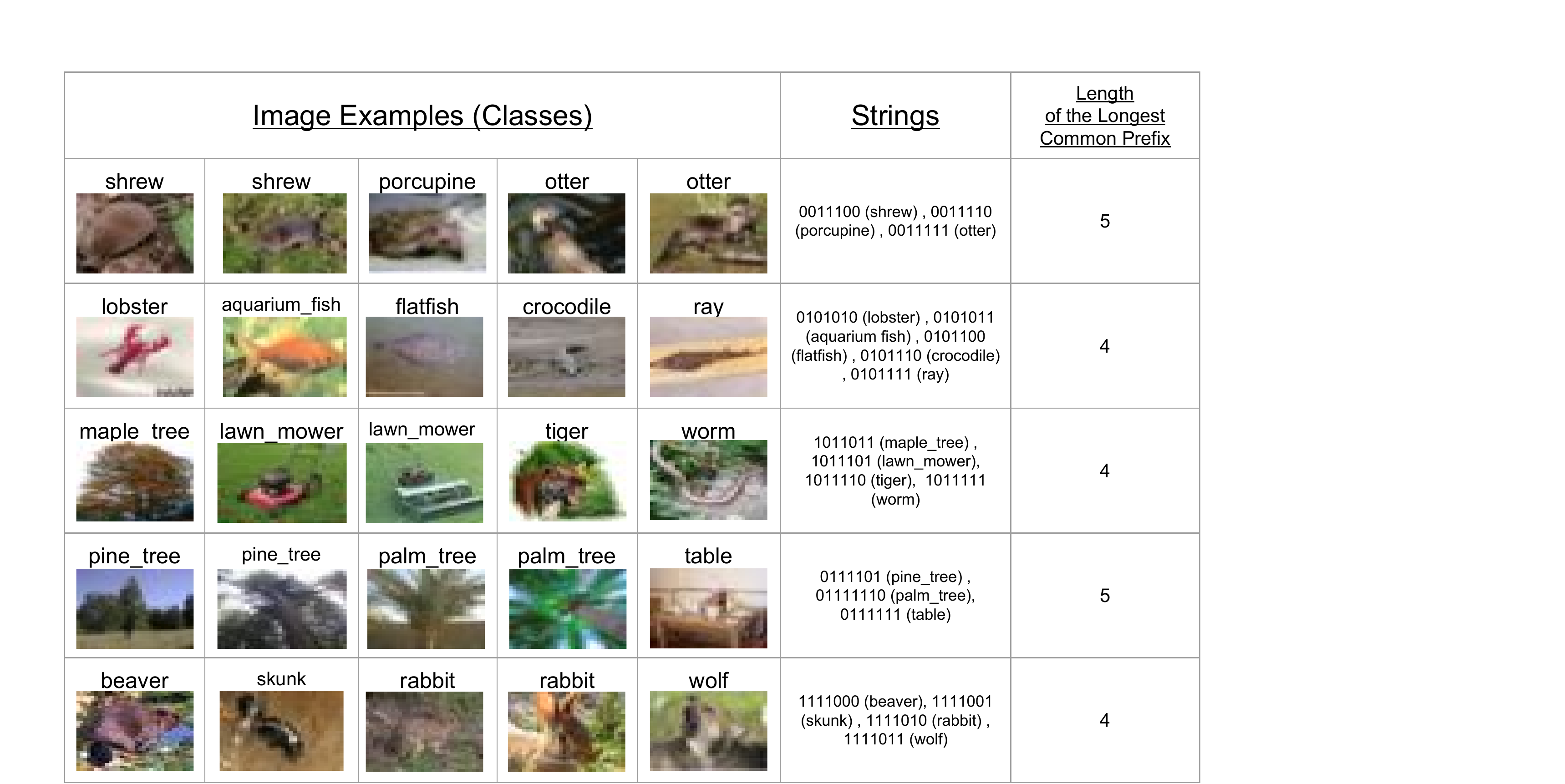}
\caption{While classes like shrew, porcupine and otter have similar visual features. Images of crocodile had water around them in the images which justifies its proximity with the other fishes, images of tiger and worm had a lot of greenery around it which has resulted in it having a long common prefix with maple tree and lawn mower. Pine tree, palm tree (with brown trunks) and table have brown as a common color while beaver, skunk, rabbit and wolf are not only visually similar but also of similar color.}
 \label{fig:visual_classes}
 \vspace{-1em}

\end{figure}

\begin{figure}[!tbh]

\centering
\includegraphics[width=7cm,height =3.8cm]{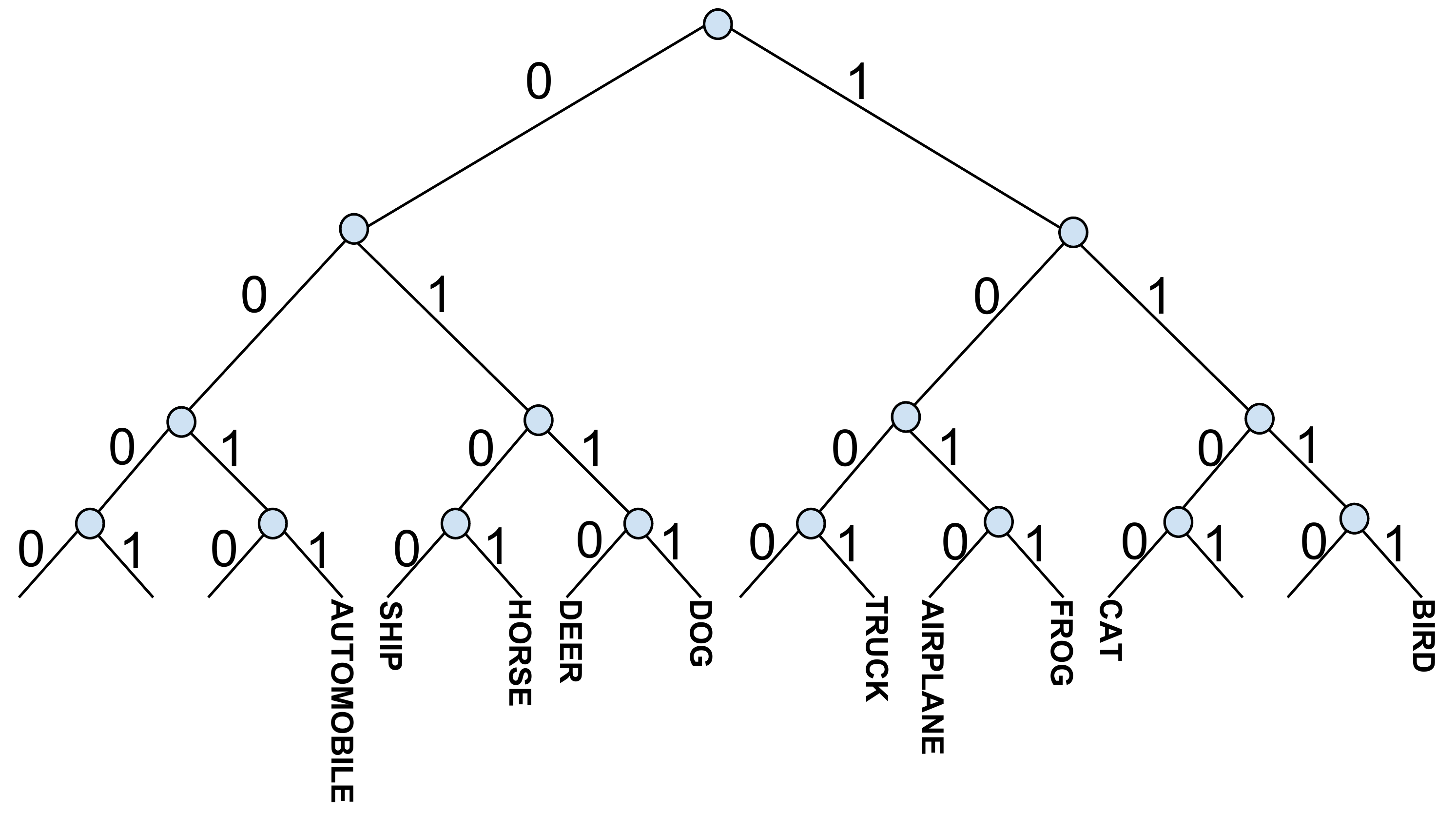}
\vspace{-1em}
\caption{Prefix tree learnt for the CIFAR10 (left) and MNIST (right) datasets.  Some of the visually similar digits in MNIST such as 3,8 and 9 have a common prefix. Hence their close proximity in the tree. Even 7 and 2 are in proximity since they are visually similar.}
\label{fig:cifar_tree}
\centering
\vspace{-1em}
\end{figure}

\textbf{Latent Hierarchy:}
In Table \ref{tab:4}, we demonstrate how our method compares with HD-CNN \cite{HDCNN} which also discovers a 2 level latent hierarchy, on CIFAR-100 and Imagenet 1K datasets. The proposed LHC is able to outperform HD-CNN in terms of the number of parameters for both the mentioned datasets while also having a lower error rate for Imagenet 1K.

Next we examine if the string embedding learnt by the Class2Str network represents a latent hierarchy.
We replace the Class2Str network by a random embedding of labels to strings that is one to one. The random embedding  is given as a look up table, which gives a string from each label. We train the LH classifier with the structured loss with the targets being the strings from the random embedding corresponding to the ground truth labels. We obtain that the accuracy reduction is 5\% for CIFAR10 and 4.3\% for CIFAR100.   

\begin{figure}[!tbh]
\centering
\includegraphics[scale=0.12]{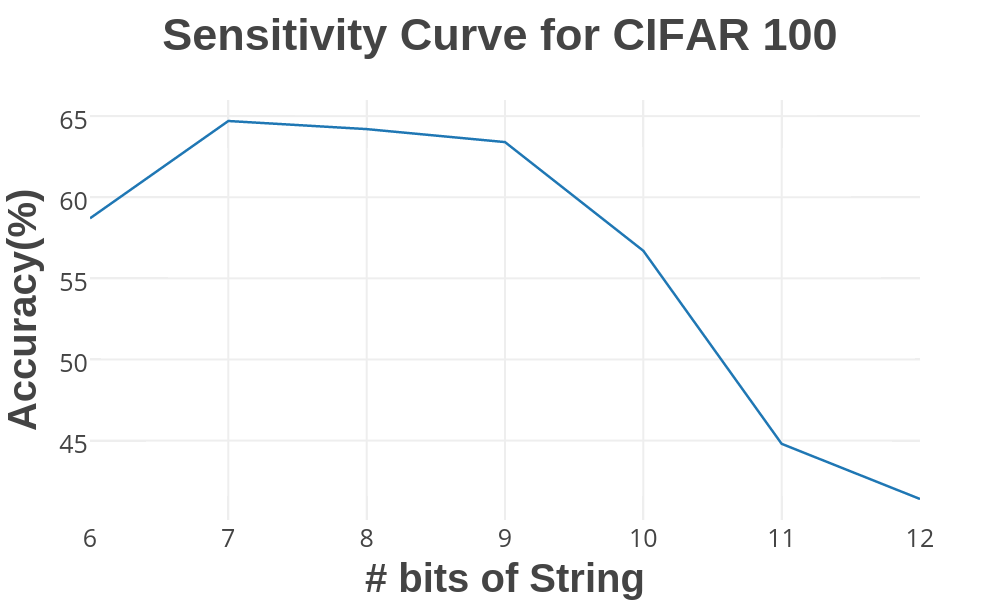}
\includegraphics[scale=0.12]{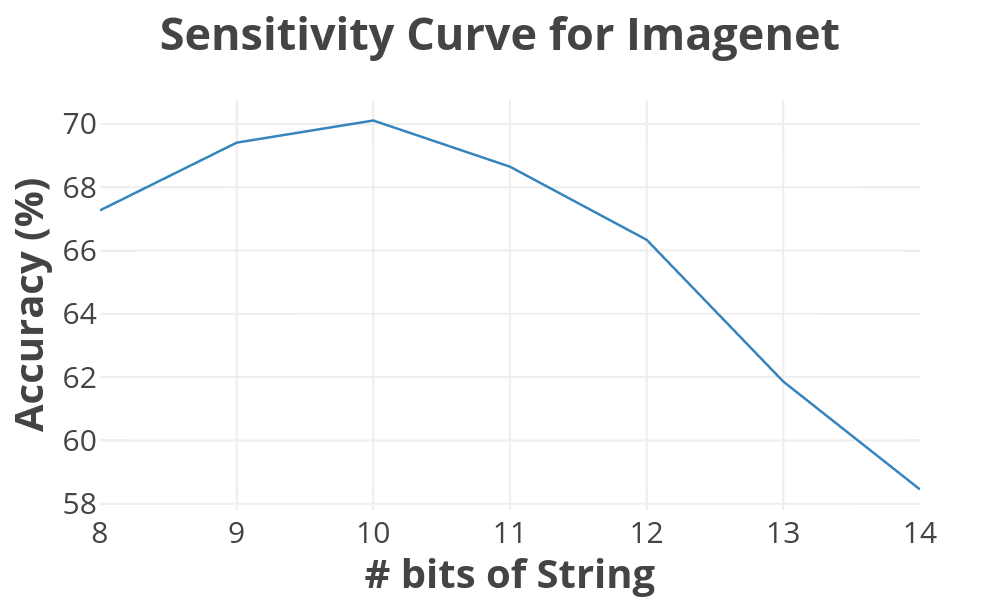}
\vspace{-1em}
\caption{Accuracy is plotted vs the the number of bits of string to be used for a latent representation of the label space, after the loss saturates and no further learning takes place. The optimal number of bits can be found from such plots for each dataset.}
 \label{fig:sensitivity}
\end{figure}

The hierarchies learnt by our models for MNIST and CIFAR10 is given in Figure \ref{fig:minst_tree}, \ref{fig:cifar_tree} respectively. As can be seen from the figure, visually similar classes seem to be grouped together. For CIFAR100, we observe some examples of visually similar classes getting mapped to strings which have a long common prefix (see Figure \ref{fig:visual_classes}). We also analyze effect of changing the values of string length $L$ on the accuracy (see Figure \ref{fig:sensitivity}). Best accuracy is obtained by choosing $L = \lceil \log_2 C \rceil$ as is evident from the plot.

\begin{figure}[!tbh]
\centering
\includegraphics[scale=0.12]{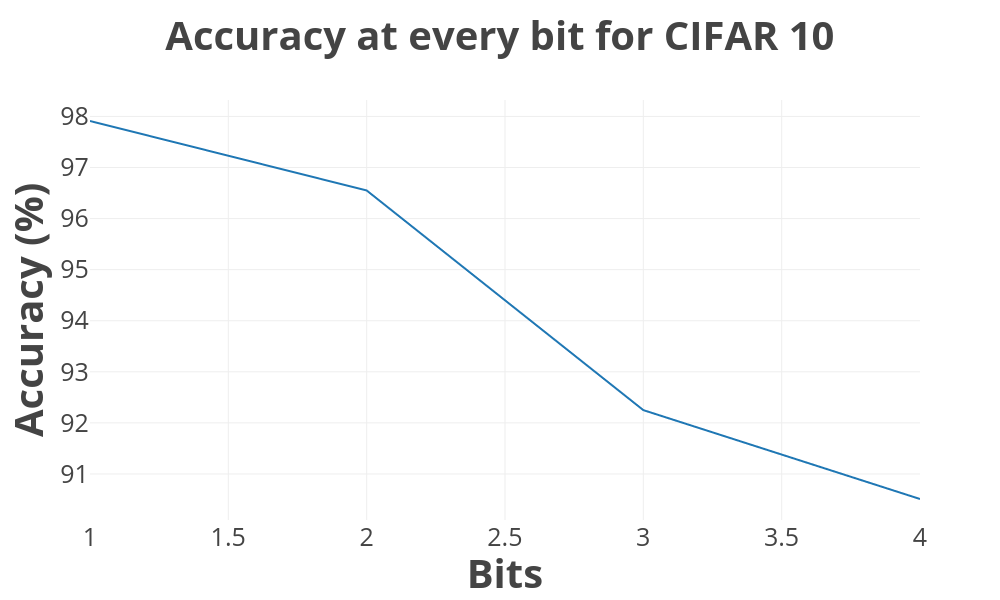}
\includegraphics[scale=0.12]{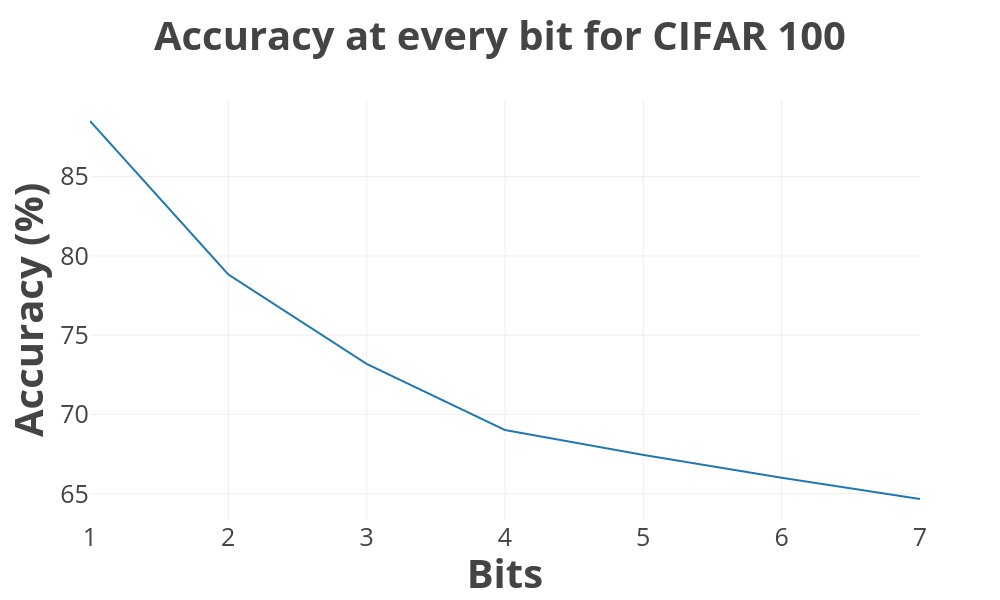}
\vspace{0.2em}
\caption{Classification Accuracy at every bit of the predicted string as a result of using Structured Loss.}
 \label{fig:bit_acc}

\end{figure}

We use a structured loss (see Section \ref{sec-loss}) during training for backpropagating the error, which assigns more penalty to misclassification at the initial bits and lower penalty to the subsequent bits. This essentially means that the network will learn to classify initial bits of the string more correctly than the later bits, and at test time we will get the best accuracy for classification of the 1st bit of the string. The accuracy reduces as we move ahead for subsequent bits (see Figure \ref{fig:bit_acc}). This seemingly natural result is verified by us by checking the accuracy at every bit position of the string the datasets. 

Note that the flipping 0's to 1's doesn't change the latent hierarchy. Hence, there are multiple Class2Str mappings corresponding to the same latent hierarchy. The system is stochastic and converges to one of them randomly. This is not necessarily a problem. However, explicit symmetry breaking in the model or training process might improve the results in the future.

\textbf{Accuracy.}
We report the accuracy of the base model with the FC Classifier as well as of the proposed model with the LH Classifier. Some of the newer architectures, replace the fully connected classifier by the global average pooling modules. Hence we compare against base models which have a FC classifier network like the Alexnet and VGG16 networks. As shown, we can recover the classification accuracy of the base model in each case (see Table \ref{tab:2}).  

\begin{figure}[!tbh]
\centering
\includegraphics[scale=0.12]{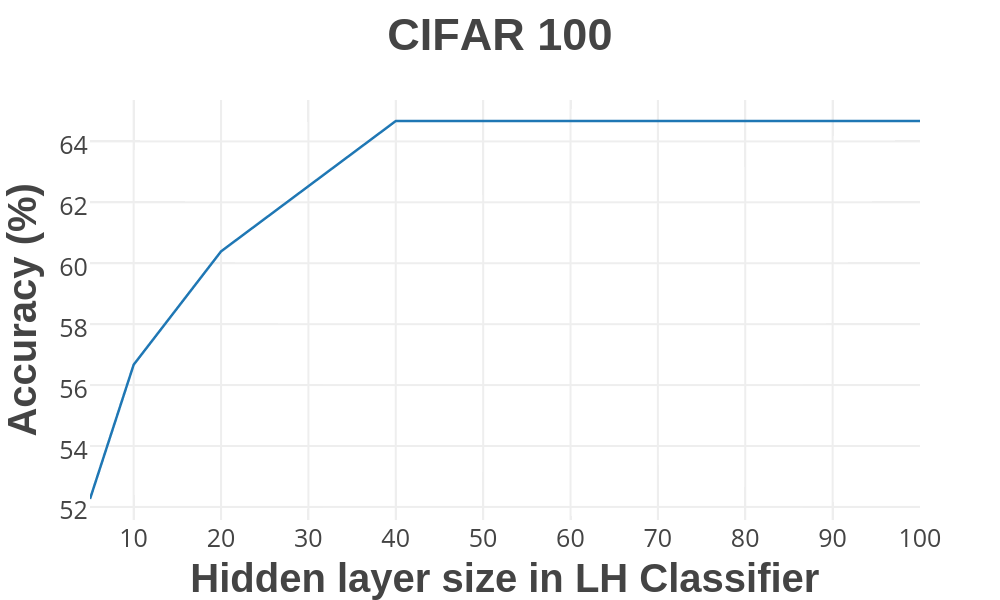}
\includegraphics[scale=0.12]{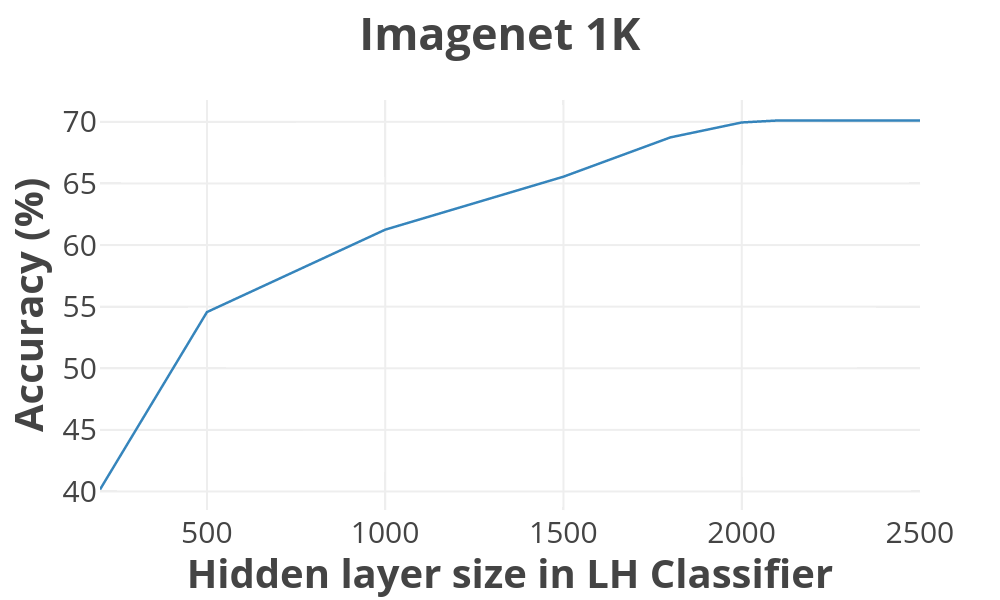}
\vspace{0.2em}
\caption{Plots for the change in accuracy with varying hidden layer sizes of the LSTM.}
\label{fig:hidden}
\vspace{-1em}
\end{figure}
\textbf{Compression.}
Our proposed method has advantages in terms of compression too. We replace a high complexity Fully Connected (FC) classifier by a low complexity LH Classifier, trained according to our proposed method. Since the Class2Str network converges to a one to one mapping (proved earlier), it can be replaced by a look up table during the testing phase. The look up table maps each string to the unique class that has been learnt. During the testing phase, the class predictions are given by first getting the string predicted by the LH classifier network, followed by a binary tree traversal. This adds negligible memory and running time to the system. Moreover, it can be seen from the table that, had we used the same number of parameters in a FC Classifier as those used by LH Classifier, we would have got significantly inferior performance thereby proving LH Classifier to be a beneficial alternative. Hence, replacing the FC Classifier with the LH classifier reduces parameters. The parameter reduction is around $96\%$ (see Table \ref{tab:2}), for CIFAR100 and around $41\%$ for Imagenet 1K dataset. 
In Figure \ref{fig:hidden}, we show how the accuracy of the model varies with different hidden state sizes of the LSTM used by the LH Classifier. As can be seen from the Figure \ref{fig:hidden}, decreasing the hidden state size to 1000, still gives a accuracy $>60\%$ for VGG16 trained on Imagenet.

\vspace{-2em}
\section{Conclusion}
In this work, we have proposed a deep neural network architecture for image classification, that explicitly learns a latent hierarchy of the classes. We have empirically demonstrated that this architecture is equally good as the traditional architectures with classifier networks in terms of accuracy. Moreover, it achieves these accuracies with a fraction of the parameter complexity, making it a compelling addition to the known model compression techniques, when applied to classifier networks. We believe it is an interesting and open research direction to use prior knowledge of hierarchies in the classes along with the LH Classifier, in order to get better accuracy. 
\pagebreak






\bibliographystyle{IEEEtran}
\bibliography{dl}


%



\end{document}